\documentclass[11pt,a4paper]{article}
\usepackage[utf8]{inputenc}
\usepackage{amsmath}
\usepackage{amsfonts}
\usepackage{amssymb}
\usepackage{graphicx}
\usepackage{hyperref}
\usepackage{inconsolata}
\usepackage{booktabs}
\usepackage{geometry}
\usepackage{cite}
\usepackage{titlesec}
\titleformat{\section}
  {\normalfont\Large\bfseries}{\thesection.}{1em}{}
\titleformat{\subsection}
  {\normalfont\large\bfseries}{\thesubsection.}{1em}{}

\hypersetup{
    colorlinks=true,
    linkcolor=blue,
    urlcolor=cyan,
    citecolor=red,
    pdftitle={Knowledge Protocol Engineering: A New Paradigm for AI in Domain-Specific Knowledge Work},
}

\title{\textbf{Knowledge Protocol Engineering: A New Paradigm for AI in Domain-Specific Knowledge Work}}
\author{
    Guangwei Zhang\thanks{Corresponding author. This work proposes a new conceptual framework for human-AI collaboration in specialized knowledge domains.} \\
    \textit{School of History and Civilization, Shaanxi Normal University} \\
    \texttt{\href{mailto:zhangguangwei@snnu.edu.cn}{zhangguangwei@snnu.edu.cn}}
}
\date{July 2025}

\begin{document}

\maketitle
\thispagestyle{empty}

\begin{abstract}
\noindent\textit{The capabilities of Large Language Models (LLMs) have opened new frontiers for interacting with complex, domain-specific knowledge. However, prevailing methods like Retrieval-Augmented Generation (RAG) and general-purpose Agentic AI, while powerful, often struggle with tasks that demand deep, procedural, and methodological reasoning inherent to expert domains. RAG provides factual context but fails to convey logical frameworks; autonomous agents can be inefficient and unpredictable without domain-specific heuristics. To bridge this gap, we introduce \textbf{Knowledge Protocol Engineering (KPE)}, a new paradigm focused on systematically translating human expert knowledge, often expressed in natural language documents, into a machine-executable \textbf{Knowledge Protocol (KP)}. KPE shifts the focus from merely augmenting LLMs with fragmented information to endowing them with a domain's intrinsic logic, operational strategies, and methodological principles. We argue that a well-engineered Knowledge Protocol allows a generalist LLM to function as a specialist, capable of decomposing abstract queries and executing complex, multi-step tasks. This position paper defines the core principles of KPE, differentiates it from related concepts, and illustrates its potential applicability across diverse fields such as law and bioinformatics, positing it as a foundational methodology for the future of human-AI collaboration.}
\end{abstract}

\vspace{1cm}

\section{Three Curves of Capability: Situating a New Paradigm for AI Specialization}

The development of Large Language Models (LLMs) can be understood as a succession of capability enhancement curves. The \textbf{first curve}, powerfully articulated by the "Scaling Laws" \cite{kaplan2020scalinglaws}, was driven by a paradigm of scale—leveraging vast computational power and web-scale data to achieve unprecedented general intelligence. While foundational, this path is now facing diminishing returns as models approach the saturation point of high-quality, general-knowledge corpora.

This saturation gave rise to the \textbf{second curve}: Retrieval-Augmented Generation (RAG) \cite{lewis2020retrieval}. This paradigm empowers end-users to enhance LLM performance by providing external, factual context. While RAG boosts accuracy on knowledge-intensive tasks, its contribution primarily lies in augmenting "static knowledge." It struggles to convey the procedural and abstract reasoning frameworks that define expert-level problem-solving. This limitation has, in turn, spurred the growth of Agentic AI, often based on frameworks like ReAct \cite{yao2022react}, which grant LLMs tools and a reasoning loop to tackle multi-step problems.

However, both RAG and general-purpose agents face challenges in specialized domains. This paper argues for the necessity and timeliness of a \textbf{third curve}, one focused not on what a model knows, but on how it thinks. We propose that this curve is defined by methodological augmentation. We introduce and define a new paradigm, Knowledge Protocol Engineering (KPE), as the foundational practice for this next stage of AI specialization. KPE aims to elevate an LLM's capabilities from factual recall to a higher order of abstract and procedural reasoning, addressing a critical gap left by the previous two curves.

\section{Knowledge Protocol Engineering (KPE)}

To address this challenge, we propose a new paradigm: \textbf{Knowledge Protocol Engineering (KPE)}. Our approach is conceptually inspired by recent industrial efforts to standardize LLM interaction, such as Anthropic's Model Context Protocol (MCP) \cite{anthropic2024mcp}. However, KPE diverges by shifting the focus from a technical, syntactic protocol to a deep, semantic one.

\textbf{Definition:} KPE is the systematic practice of designing and refining human expert knowledge—codified in documents like manuals, academic frameworks, or standard operating procedures—into a machine-executable Knowledge Protocol (KP). This protocol's purpose is to guide and constrain an LLM's reasoning and operational behavior within a specific, complex domain.

KPE is founded on a core philosophical shift: instead of treating human-authored documents as a passive corpus for retrieval, we treat them as the source code for an active, structured protocol. The role of the human expert is elevated from a user to a \textbf{Knowledge Architect} or \textbf{AI Mentor}.

KPE operates on three core principles:
\begin{enumerate}
    \item \textbf{Methodology as First-Class Citizen:} The protocol's primary payload is not information, but methodology. It encodes workflows, decision trees, logical dependencies, and heuristic strategies.
    \item \textbf{Human-Centric Authoring:} The source of a KP is a human-readable document. This elevates the role of the domain expert, transforming their writing and structuring of knowledge into a direct form of AI programming.
    \item \textbf{Holistic Contextualization:} Unlike the fragmented nature of RAG, a KP aims to provide a coherent, logically-connected ``mental model" or ``worldview" of the domain, enabling the LLM to understand relationships between concepts and procedures.
\end{enumerate}

\section{Illustrative Use Cases}

To demonstrate the potential of KPE, we present two hypothetical examples from different domains.

\subsection{Case 1: Legal Analysis}
Consider a legal database containing corporate law statutes. A junior associate asks: ``Does Company X's proposed merger with Company Y violate anti-monopoly regulations, considering their respective market shares in the software and hardware sectors?"

\begin{itemize}
    \item \textbf{Standard RAG approach}: It would retrieve general articles about anti-monopoly law and perhaps the definition of ``market share." It would likely fail to construct a valid legal argument.
    \item \textbf{KPE approach}: A Knowledge Protocol would be engineered from a treatise on antitrust law. It would contain a \textbf{strategy} like:
    \begin{enumerate}
        \item \textit{Define relevant market}: A procedure to query for market definitions based on product sectors (software, hardware).
        \item \textit{Calculate HHI}: A rule to calculate market concentration using the Herfindahl-Hirschman Index.
        \item \textit{Apply safe harbor rules}: A decision tree based on the calculated HHI to determine if the merger falls into a ``safe harbor" category.
    \end{enumerate}
    Guided by this protocol, an LLM could generate a sequence of queries and logical steps to produce a structured, methodologically sound preliminary analysis.
\end{itemize}

\subsection{Case 2: Bioinformatics Workflow}
A biologist wants to: ``Find all genes in the human genome that are associated with Alzheimer's disease and are also targeted by the experimental drug `Pharma-X'."

\begin{itemize}
    \item \textbf{Standard Agentic approach}: An agent might try to query a gene database for ``Alzheimer's" and a chemical database for ``Pharma-X" separately. It might struggle to formulate a plan to correctly intersect these two complex datasets.
    \item \textbf{KPE approach}: A Knowledge Protocol is built from a standard bioinformatics pipeline document. It specifies the \textbf{workflow}:
    \begin{enumerate}
        \item \textit{Query Disease-Gene Association}: Use the `GeneDB' to find all genes linked to ``Alzheimer's disease" (Concept ID: `C0002395').
        \item \textit{Query Drug-Target Association}: Use the `DrugBank' to find all gene targets of ``Pharma-X" (DBID: `DB12345').
        \item \textit{Intersect Results}: Perform a `JOIN' or set intersection on the results from the previous two steps to find the common genes.
    \end{enumerate}
    The protocol provides the LLM with a clear, unambiguous, and validated scientific workflow, transforming a complex query into a series of executable steps.
\end{itemize}

\section{Differentiating KPE: A Paradigm for Methodological Augmentation}

KPE's uniqueness is best understood by contrasting its core tenets with existing paradigms. While it belongs to the broader field of Context Engineering, its philosophy and mechanism constitute a distinct approach (see Table \ref{tab:comparison}).

\begin{table}[h!]
\centering
\caption{KPE in the Landscape of AI Paradigms}
\label{tab:comparison}
\resizebox{\textwidth}{!}{%
\begin{tabular}{@{}llll@{}}
\toprule
\textbf{Paradigm} & \textbf{Core Function} & \textbf{Knowledge Unit} & \textbf{Metaphor} \\ \midrule
\textbf{Standard RAG} & Fact Augmentation & Data Chunks & Open-Book Examinee \\
\textbf{Agentic RAG} & Fact-Augmented Action & Data Chunks & Detective with a Phone \\
\textbf{Context Engineering} & General Context Optimization & Any & Toolbox User \\
\textbf{Knowledge Protocol Eng. (KPE)} & \textbf{Methodology Injection} & \textbf{Logical Protocol Blocks} & \textbf{Apprentice with a Manual} \\ \bottomrule
\end{tabular}%
}
\end{table}

The critical distinction lies in the nature of the knowledge being transferred. While prevailing paradigms focus on augmenting LLMs with static, factual knowledge, KPE is engineered to convey dynamic, procedural, and abstract knowledge.

\begin{itemize}
    \item \textbf{From Information to Methodology (KPE vs. RAG):} Standard RAG excels at augmenting LLMs with declarative, ``what-is" knowledge that can be represented in vector space. It answers the question, ``What does the model need to know?". KPE, conversely, answers a fundamentally different question: ``How must the model \textit{think and act}?". While RAG provides factual nouns (e.g., the content of a database table), KPE provides operational verbs and logical adverbs in the form of an executable procedure (e.g., the \textit{method} for joining specific tables to analyze an abstract concept). This allows a KPE-driven system to tackle problems of a significantly higher order of complexity.

    \item \textbf{From Retrieving Facts to Retrieving Methods (KPE vs. Agentic RAG):} This distinction becomes even sharper when considering agentic systems. An Agentic RAG agent uses retrieval to fill factual gaps in its world model (e.g., ``What is the protein target of this drug?"). A KPE-driven agent, on the other hand, uses retrieval to fill methodological gaps in its plan (e.g., ``What is the established academic procedure for analyzing `social mobility'?"). The former retrieves data to inform its next step; the latter retrieves a strategy to structure its entire plan.

    \item \textbf{From Toolbox to Architecture (KPE vs. Context Engineering):} Context Engineering represents the broad ``toolbox" of all techniques for optimizing prompts. KPE is a specific ``architectural philosophy" derived from that toolbox. It posits that for any deep knowledge domain, the optimal context is not an ad-hoc collection of facts or examples, but a holistic, human-authored protocol that codifies the domain's expert methodologies.
\end{itemize}

In essence, KPE is a specialized discipline that moves beyond retrieving what an expert knows to codifying how an expert thinks, thereby augmenting the model's \textbf{reasoning architecture}, not just its knowledge base.

\section{Conclusion: KPE as the Post-Training Paradigm for the Third Curve of AI}

The trajectory of AI development beckons us toward a new frontier. As the ``first curve" of scaling generalist models matures, and the ``second curve" of fact-based RAG reveals its limitations in handling procedural complexity, we stand at the cusp of a \textbf{third curve}—one defined not by the breadth of knowledge, but by the depth of methodology. Knowledge Protocol Engineering (KPE) is, we argue, the foundational paradigm for this new era.

KPE offers more than just a technique; it proposes a fundamental shift in how we approach AI specialization, ushering in what we term the ``Post-Training" phase of development. In this phase, the primary driver of value creation is no longer the costly and data-intensive process of altering a model's internal weights. Instead, value is generated by specializing generalist models through high-quality, targeted, and context-injected guidance. KPE is the art and science of this guidance.

This new paradigm redefines the roles of both human and machine. It conceptualizes the human domain expert not as a passive user, but as a ``teacher" or ``Knowledge Architect" who designs the curriculum for an AI apprentice. It envisions a future where our most valuable human legacy—the complex, procedural, and methodological wisdom of our respective fields—is not simply archived for human consumption, but actively engineered into ``thought-ware" that can direct and empower AI. Just as human societies rely on professional education and apprenticeship to transform bright individuals into experts, KPE provides a scalable and efficient pathway to ``specialize" generalist AIs.

In conclusion, KPE responds to a critical need in the contemporary AI landscape. It provides a robust framework for elevating LLMs from knowledgeable generalists to methodologically sound specialists. By placing human expertise at the very center of AI augmentation, KPE represents a sustainable, collaborative, and profoundly humanistic future for the development of artificial intelligence in all fields of knowledge work.

\end{document}